\documentclass[10pt,twocolumn,letterpaper]{article}

\usepackage{cvpr}              

\usepackage{graphicx}
\usepackage{enumitem}
\usepackage{amsmath}
\usepackage{amssymb}
\usepackage{booktabs}
\usepackage{microtype}
\usepackage{balance}
\usepackage{subcaption}
\usepackage{multirow}
\usepackage[ruled]{algorithm2e}
\newcommand{\redbf}[1]{{\textbf{\color{red}{#1}}}} 
\newcommand{\blueud}[1]{{\underline{\color{blue}{#1}}}} 
\makeatletter
\newcommand{\printfnsymbol}[1]{%
	\textsuperscript{\@fnsymbol{#1}}%
}
\makeatother

\usepackage{hyperref}
\hypersetup{pagebackref,breaklinks,colorlinks}

\usepackage[accsupp]{axessibility}
\usepackage[capitalize]{cleveref}
\crefname{section}{Sec.}{Secs.}
\Crefname{section}{Section}{Sections}
\Crefname{table}{Table}{Tables}
\crefname{table}{Tab.}{Tabs.}

\def\cvprPaperID{4335} 
\def\confName{CVPR}
\def\confYear{2022}

\begin{document}

\title{Escaping Data Scarcity for High-Resolution Heterogeneous Face Hallucination}

\author{Yiqun Mei\thanks{equal contribution} , Pengfei Guo\printfnsymbol{1}, Vishal M. Patel \\{Johns Hopkins University}}
\maketitle

\begin{abstract}
In Heterogeneous Face Recognition (HFR), the objective is to match faces across two different domains such as visible and thermal. Large domain discrepancy makes HFR a difficult problem. Recent methods attempting to fill the gap via synthesis have achieved promising results, but their performance is still limited by the scarcity of paired training data. In practice, large-scale heterogeneous face data are often inaccessible due to the high cost of acquisition and annotation process as well as privacy regulations. In this paper, we propose a new face hallucination paradigm for HFR, which not only enables data-efficient synthesis but also allows to scale up model training without breaking any privacy policy. Unlike existing methods that learn face synthesis entirely from scratch, our approach is particularly designed to take advantage of rich and diverse facial priors from visible domain for more faithful hallucination. On the other hand, large-scale training is enabled by introducing a new federated learning scheme to allow institution-wise collaborations while avoiding explicit data sharing. Extensive experiments demonstrate the advantages of our approach in tackling HFR under current data limitations. In a unified framework, our method yields the state-of-the-art hallucination results on multiple HFR datasets.   
\end{abstract}


\section{Introduction} \label{sec:1}
Deep convolutional neural networks have led to unprecedented success on visual face recognition~\cite{wen2016discriminative,schroff2015facenet,deng2019arcface,wang2018cosface}, where state-of-the-art methods achieve more than 99\% accuracy on multiple benchmarks. These near-perfect performances come from both well-elaborated architectures and exhaustive training on massive datasets.  Nevertheless, in many real-world scenarios with low-visibility, such as low-light and night-time, it is often infeasible to obtain clear visible (VIS) images. Under these circumstances, sensors deployed for other imaging spectra, e.g. Thermal (TH), can capture more discriminative information and serve as a more reliable solution. This raises a great need of heterogeneous face recognition (HFR)~\cite{liu2012heterogeneous,li2007illumination,li2009hfb}, an important task in computer vision and biometrics, that matches images from TH modality to its VIS counterpart \footnote{Note that HFR is a general term that is used for matching two face images taken in two different domains such as TH, VIS or sketch.  In this paper, we refer to HFR as a specific problem of matching TH images with VIS images.}.  The HFR problem has numerous applications in surveillance, monitoring and security. 

Unfortunately, due to the large domain discrepancy, naively deploying a state-of-the-art face recognition algorithm trained on VIS images often leads to poor performance on a TH dataset~\cite{he2018wasserstein}. Over the past decade, tremendous efforts have been spent to address the HFR problem by either learning domain-invariant features~\cite{liao2009heterogeneous,galoogahi2012face,gong2017heterogeneous,fu2021cm} or finding a common subspace~\cite{yi2007face,klare2012heterogeneous,shao2014generalized,peng2020soft}.  Owing to the rapid progress in Generative Adversarial Networks (GANs)~\cite{goodfellow2014generative}, most recent methods~\cite{fu2019dual,ijcai2019-143,duan2020cross,song2018adversarial,zhang2017generative,di2018polarimetric} reformulate HFR as a face synthesis/translation problem. The resulting ``recognition via hallucination" scheme embraces a huge benefit that any off-the-shelf recognizer can be directly applied on the synthesised images. 

While these synthesis-based approaches fill the gap to some extent, the produced VIS images are still unsatisfactory, often accompanied with distorted and incorrect facial structures (shown in Figure \ref{vpgan_fig}), which significantly degrades the recognition accuracy. We found that the bottleneck is likely due to the limited size of the dataset which fails to offer sufficient information to guide image synthesis. Unlike visible images that are easy to obtain and widely available over the Internet~\cite{guo2016ms}, collecting and annotating a large-scale high-quality TH dataset is difficult. Challenges stem from many aspects.  First, the acquisition process is both time-consuming and costly, which often requires laborious setup and non-trivial calibrations~\cite{poster2021large,wang2008face}. Second, the diversity of collected data can be limited. Due to the physical constraints, it is typically infeasible for a single institution to collect a comprehensive dataset that covers a diverse set of identities with various attributes, such as race, gender and age. Most existing datasets~\cite{mallat2018benchmark,panetta2018comprehensive,huang2012buaa,espinosa2013new} are confined to a small number of subjects, leading to biased results and over-fitting. Third, face data is privacy sensitive. Since it contains subjects' personal identification information, one has to deal with privacy concerns when collecting and storing them, making it difficult to share the data with other institutions. 

Besides poor synthesis quality, most existing methods can only process images at a resolution no more than $128 \times 128$. This not only leads to visually unappealing results, but also reduces their applicability in many downstream tasks that depend on high-resolution inputs, such as face parsing~\cite{lin2019face,liu2017face}, editing~\cite{zeng2021sketchedit} and reenactment~\cite{Pumarola_2018_ECCV}. 

In this paper, we present a unified hallucination framework for HFR, that is capable of synthesizing high-resolution visible faces ($512\times512$) from low-resolution heterogeneous data (i.e. smaller than $128\times128$), with superior realness and higher fidelity. Our approach consists of two separate strategies. The first one comes with a new generation paradigm inspired by the recent success in GAN inversion~\cite{xia2021gan}. The core idea is to leverage rich and diverse facial priors from the visible domain to eschew the need of learning generation from scratch. This is accomplished by embedding a pre-trained GAN (e.g. StyleGAN~\cite{karras2020analyzing,karras2019style}) as a facial decoder which hallucinates visible faces conditioned on the latent representations of a U-shaped encoder. The encoder is carefully designed with a novel Multi-scale Contexts Aggregation (MSCA) mechanism which merges scale-wise information to enhance representation. MSCA offers better fine-grained generation control and is pivotal for preserving identity information.  The proposed method, called \textit{\textbf{V}isual}  \textit{\textbf{P}rior enhanced \textbf{GAN} (VPGAN)}, can break the underlying data limitation by producing faces with state-of-the-art accuracy and photo-realism. 

Deep models are data hungry and VPGAN may be further improved with large-scale training. However, in practice, HFR data tends to be separately collected and dispersed among different intuitions. Due to privacy concerns, one cannot simply share the data for centralized training. To this end, our second strategy introduces Federated Learning (FL)~\cite{konevcny2016federated} to further improve HFR, which enables collaborative model training while avoiding explicit data exchanging. Specifically, we allow each institution to perform local training on their private HFR data and deploy a centralized server to periodically communicate with each institution, aggregating local models and updating a global model. The whole process does not involve any data transfer but benefits deep models a lot by integrating information from a significantly broader range of data. To make our approach more suitable for real-world HFR, we have to tackle the heterogeneous data distributions across institutions. This challenge, which may be caused by differences in sensor types or acquisition protocols, can lead to locally skewed updates, resulting in slow convergence and sub-optimal performance~\cite{li2019convergence,li2021federated}. To tackle this issue, we build our FL algorithm based on a new Model Proximity Regularization (MPR), which corrects local gradient updates by constraining the discrepancy between the latent representations from the global and local models. As a result, our approach achieves superior robustness towards non-ideal data distribution, implying its applicability for solving real-world HFR problems.

In our unified framework, we integrate VPGAN as the basic component and use it in the proposed FL scheme. We term this new framework VPFL. In the experiment section, we demonstrate that our approach generates VIS faces at high-resolution with superior realness and accuracy.

Within the infrared spectrum, various modalities have been explored for thermal-to-visible (TH-VIS) face recognition.  These include Near Infrared (NIR), Short-Wave Infrared (SWIR), Mid-Wave Infrared (MWIR) and Long-Wave Infrared (LWIR). The use of a particular thermal modality depends on the application.  For instance, in long-range surveillance applications, SWIR or MWIR modalities are often used.  Unlike NIR images, which are close to the VIS spectrum images, LWIR images are often acquired in low-resolution with many facial details missing on the captured imagery. This is well-reflected by the performance drop of the existing recognition methods on such datasets \cite{immidisetti2021simultaneous, HFR_FG21}.  Also, greater than 99\% accuracy has been reported on many NIR face datasets \cite{duan2020cross,fu2019dual}. However, the performance of various HFR methods on LWIR data is significantly low \cite{immidisetti2021simultaneous, HFR_FG21}.

In summary, the main contributions of our paper are: 
\begin{itemize}[noitemsep]
    \item We propose a new data-efficient generation scheme for HFR. Thanks to the powerful visual priors, it manages to alleviate fundamental data challenges, resulting in superior hallucination results.
    \item We introduce a unified framework to make large-scale training possible in real-world scenarios. VPFL makes multi-institutional collaborations possible without raising any privacy concerns.
    \item Extensive experiments show that our method can produce faces with state-of-the-art photo-realism and fidelity, which in turn significantly boosts the recognition accuracy. These merits show its great potential to serve as a universal solution towards real-world applications.
\end{itemize}

\section{Related Work}
\begin{figure*}[t] 
\centering
\includegraphics[clip, trim=0cm 4.88cm 0cm 0cm, width=0.9\textwidth ]{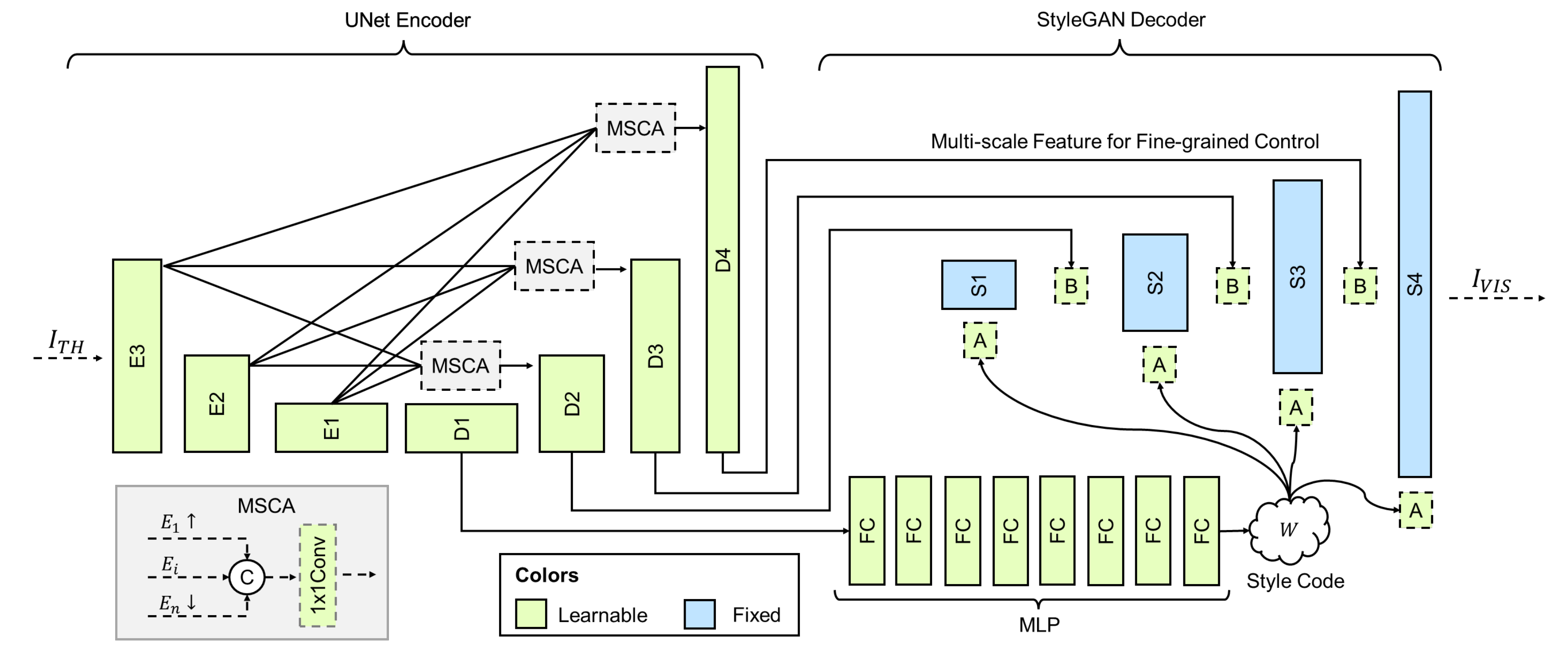}
\vspace{-2.6mm}
\caption{The proposed VPGAN. It adopts an encoder-decoder structure. The UNet encoder extracts style codes as well as multi-scale representations and then transmits them to the decoder for generation control. MSCA enhances the encoder by merging multi-scale information, which proves to be crucial for accurate hallucination. A pre-trained StyleGAN~\cite{karras2019style,karras2020analyzing} serves as the facial decoder and generates the desired visible face. A is the linear transformation of the style codes in~\cite{karras2019style,karras2020analyzing} and B plays a similar role  as that of \textit{noise injection}.}
\label{gan}
\vspace{-6mm}
\end{figure*}

\noindent\textbf{Heterogeneous Face Recognition.} Compared to the methods relying on feature/subspace learning,  recognition via synthesis~\cite{duan2020cross,mallat2019cross,zhang2017generative,zhang2019synthesis,fu2019dual,fu2021dvg,fu2021high} has received significant attention, because off-the-shelf face recognition algorithms can be applied to the synthesized images. Our discussion will focus on these types of methods. Early deep learning-based approaches directly learn a CNN for cross-spectral mapping. For example, Lezama \etal~\cite{lezama2017not} train a CNN for NIR-VIS and improve results with low-rank assumption. Riggan \etal~\cite{riggan2018thermal} enhance the
discriminative quality of the synthesized images by modeling both global and local regions. Recent methods leverage GANs to improve the hallucination qualities. Zhang \etal~\cite{zhang2019synthesis} propose GAN-VFS that jointly learns semantic rich features and facial reconstruction, which results in more photo-realistic and accurate generation. Further improvements are based on cycle-consistency~\cite{song2018adversarial}, more advanced loss~\cite{di2018polarimetric}, and attention mechanism~\cite{immidisetti2021simultaneous,di2019polarimetric,ijcai2019-143}. 

\noindent\textbf{GAN Inversion.}
Our method is related to GAN inversion~\cite{menon2020pulse,wang2021towards,yang2021gan,chan2021glean,gu2020image,richardson2021encoding} which relies on pre-trained GAN priors for better image manipulation and restoration. Early approaches~\cite{menon2020pulse,gu2020image} explicitly ``invert GANs'', which iteratively find the closest latent code of a targeting image. For example, PULSE~\cite{menon2020pulse} for photo upsampling gradually searches the correct latent code of a StyleGAN~\cite{karras2019style,karras2020analyzing} by optimizing a downsampling loss. More recent methods~\cite{wang2021towards,yang2021gan,chan2021glean,richardson2021encoding} use a DNN encoder and learn to predict the latent code in one forward pass. Our work is inspired by these approaches but exploits its ability in transferring visual priors for data-efficient heterogeneous face hallucination.

\noindent\textbf{Federated Learning.}
Federated learning is a decentralized machine learning framework which leverages data from multiple institutions or users to collaboratively train a global model without directly sharing their local data. Addressing heterogeneous data distribution across devices or institutions in the real-world deployment of FL applications draws emerging attention. Several FL methods~\cite{fedopt,afl, feddyn, fedbn,flmr,fedprox, guo2022auto, xu2022closing} targeting on this issue are built upon FedAvg~\cite{fedavg}. FedProx~\cite{fedprox} and Agnostic Federated Learning (AFL)~\cite{afl} introduce an additional regularization on weights during the local training to alleviate the learning bias issue of the global model. FedDyn~\cite{feddyn} is proposed to address the issue that there is an inconsistency between minima of the local model loss and those of the global loss by introducing a dynamic regularizer in each client. While those works conduct rigorous theoretical analysis, their performance is not validated on practical applications. Recently, Aggarwal \etal~\cite{fedface} proposed face recognition methods based on the FL framework. However, it is worth noting that the multi-institutional collaborative approach based on FL for face hallucination has not been well studied in the literature.
\section{Proposed Method}

In this section, we first formulate the TH-VIS face hallucination problem and then describe our method in detail.  Given a TH image $\mathcal{I_{TH}}$, our goal is to reconstruct a VIS face $\mathcal{I_{VIS}}$ by learning a mapping function $\mathcal{I_{VIS}} = \mathcal{F(I_{TH})}$. The synthesized face $\mathcal{I_{VIS}}$ should be both visually realistic and accurate, and thus can be used for face matching. As discussed earlier, due to various reasons, $\mathcal{I_{TH}}$ is often captured in low-resolution. Unlike existing methods (which only synthesize faces at $128\times 128$), our work performs joint translation and upsampling. To the best of our knowledge, this is the first work that can hallucinate high-resolution faces ($512\times512$) in HFR. 
\subsection{VPGAN}
 Conventional methods learn synthesis entirely from  paired datasets. Due to data limitations, they can hardly output clear and high-quality images. Our method instead only learns to control generation by making use of diverse visual priors encapsulated in a pre-trained GAN. We leverage off-the-shelf StyleGAN~\cite{karras2019style,karras2020analyzing} pretrained on FFHQ~\cite{karras2019style}, which contains 70,000 high-resolution faces. As shown in Figure \ref{gan}, VPGAN adopts an encoder-decoder architecture. Only the encoder is required to train with the HFR dataset for guiding the hallucination. To output a visible face, we first extract global style codes 
 \setlength{\belowdisplayskip}{2pt} \setlength{\belowdisplayshortskip}{2pt}
\setlength{\abovedisplayskip}{2pt} \setlength{\abovedisplayshortskip}{2pt}
\begin{align}
w= MLP(UNet_{E}(\mathcal{I_{TH}})),
\end{align}
where $\mathcal{U}Net_{E}$ and MLP are the encoder part of the UNet and fully connected layers, respectively. Then we compute a set of multi-scale features $F_{i}$ from every stage of the decoder for fine-grained generation control, \ie $F_{i} = \mathcal{U}Net_{D_{i}}(I_{TH})$. A visible face can then be produced via 
\begin{align}
\mathcal{I_{VIS}} = \mathcal{S}(w,\{F_{1},...,F_{n}\}),    
\end{align}
where $\mathcal{S}$ is the StyleGAN decoder. Thanks to diverse visual priors such as face geometry, color and texture, VPGAN is able to alleviate the need of large datasets and yields more faithful results.
\\
\noindent\textbf{Improved UNet Encoder.}
U-shaped structure has shown great capability in obtaining semantic-rich representations. However, we found that a naive UNet~\cite{UNet} is incapable of generating accurate local structures, which are crucial for face-matching. This is because features in the decoder are upsampled from coarser scales, and thus lack sufficient fine-grained information. Errors from early stages are also transmitted to the subsequent layers, leading to incorrect synthesis. To this end, we improve the conventional UNet~\cite{UNet} with a new Multi-Scale Contexts Aggregation (MSCA) mechanism, which provides a comprehensive encoding of the input image at
multiple scales. This ensures the resulting features at all levels contain both coarse and fine information and thus leads to better generation control. As shown in Figure \ref{al:1}, MSCA computes an output by adaptively merging multi-scale features from the UNet encoder (after first up-scaling and down-scaling to match the spatial dimensions). Formally, the output features at $i$-th level can be expressed as follows:
 \begin{align}
     {out}_{i} = MSCA(E_{1}\uparrow,...,E_{i},...,E_{n}\downarrow),
 \end{align}
 where $\uparrow$ and $\downarrow$ denote the upscaling and downscaling operation, respectively.  
 We will demonstrate that this simple design is crucial for accurate face matching in Section~\ref{sec:ab}.\\
\noindent\textbf{Embedded Visual Prior.}
Our key design is to utilize visual priors embedded in a pre-trained GAN. The StyleGAN~\cite{karras2019style,karras2020analyzing} decoder stores diverse facial knowledge and acts similar to a memory bank or dictionary, where the extracted style codes (from encoder) query desired faces. The style codes $w$ can be incorporated either in a similar way to~\cite{karras2020analyzing}, by directly applying the modulation and demodulation operations on the convolution kernel of each style block, or through AdaIN~\cite{huang2017arbitrary} as used in~\cite{karras2019style}. Either operation is easy to implement and results in good generation quality. The multi-scale features from the UNet act in a similar way as the \textit{noise injection} in the original StyleGAN. But rather than achieving localized variation, here we want to control detailed facial components to be consistent with the input image. We achieve this via a modulation operation (B in Figure \ref{gan}) similar to ~\cite{wang2018recovering,park2019semantic}. Specifically, we compute pixel-wise scale and shift parameters $\gamma_{i}$ and $\beta_{i}$ from the $i$-th multi-scale feature $F_{i}$ via a simple $1\times 1$ convolution layer. And then we use it to calibrate the output feature from the $i$-th StyleGAN layer:
\begin{align}
    S^{+}_{i} = \gamma_{i} \odot{S^{-}_{i}}+\beta_{i},
\end{align}
where $\odot$ denotes Hadamard product and $S^{-}_{i}$ is the pre-modulated feature. After calibration, the resulting $S^{+}_{i}$ is passed to the next stage for subsequent generation. \\
\noindent\textbf{Training Objectives.}
To ensure both realness and fidelity, our training objective consists of four terms: (1) reconstruction loss $L_{r}$, (2) adversarial loss $L_{adv}$, (3) perceptual loss $L_{p}$, and (4) Identity Loss $L_{id}$. The overall loss can be expressed as follows:
\begin{align}\label{gen}
    L_{gen} = L_{r} + \lambda_{a} L_{adv} + \lambda_{b} L_{p} + \lambda_{c} L_{id},
\end{align}
where $\lambda_{a}, \lambda_{b}$ and $\lambda_{c}$ are corresponding balancing parameters. We define the reconstruction loss as the standard $L_{1}$ distance between the synthesized image $\mathcal{I_{VIS}}$ and ground-truth image $\mathcal{I_{GT}}$ to ensure content consistency. The adversarial loss is directly inherited from StyleGAN~\cite{karras2019style,karras2020analyzing} for more sharp generation.
To improve visual quality while preserving identity, we further adopt the perceptual loss and the identity loss. Both can be expressed as feature-wise distance of a given CNN (\eg, a pre-trained VGG):
\begin{align}
    L_{p}, L_{id} = \frac{1}{H_{i}W_{i}C_{i}}\|V_{i}(\mathcal{I_{GT}})-V_{i}(\mathcal{I_{VIS}})\|_{1},
\end{align}
where $V$ is the corresponding CNN. $H_{i}$,$W_{i}$,$C_{i}$ are the height, width and channel number of the $i$-th feature map in $V$. Here, we use an ImageNet pre-trained VGG for $L_{p}$ and a simple ArcFace~\cite{deng2019arcface} for $L_{id}$.

\subsection{Face Hallucination with Federated Learning}
Even with an advanced design like VPGAN, training with limited data is still an essential shortcoming for deep models. To this end, we consider Federated Learning (FL) and introduce a novel Model Proximity Regularization (MPR). In this section, we will first describe a vanilla FL framework and then detail the proposed MPR. 
\\
\begin{figure}[t!]
	\centering
	\includegraphics[clip, trim=0cm 0.0cm 16.2cm 0cm, width=0.95\columnwidth]{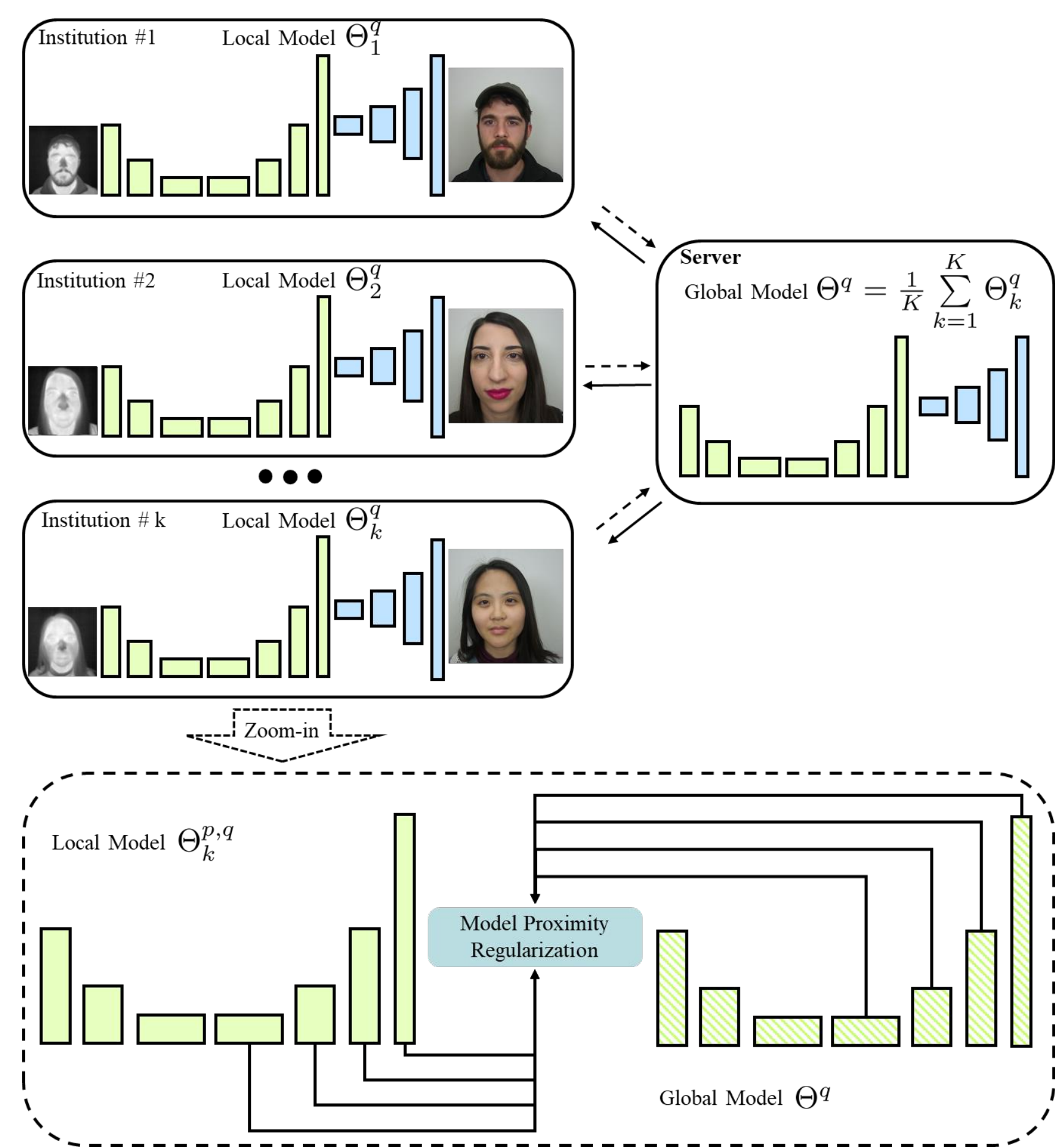}
\vskip-8pt	\caption{An overview of the proposed FL framework. Through $q$ rounds of communication between data centers and the server, the collaboratively trained global model parameterized by $\Theta^q$ can be obtained in a data privacy-preserving manner. The zoomed-in subplot shows the proposed Model Proximity Regularization (MPR) in a local institution $k$.}\label{fig2}
\vspace{-4.5mm}
\end{figure}
\noindent\textbf{A Vanilla FL Framework.} We start by recalling issues in the standard local training. Suppose we have $K$ HFR datasets $\mathcal{D}^1,\mathcal{D}^2, \dots, \mathcal{D}^K$ dispersed at different institutions. These institutions can be different universities, government agencies or private companies. In a conventional non-collaboration scheme, a local model (parameterized by $\Theta_{^k}$) at institution $k$ is trained with its own private data $D^{k}$, by optimizing $L_{gen,k}$ defined in Eq.~\ref{gen}:
\begin{equation}\Theta_{k}^{p+1} \leftarrow \Theta_{k}^{p}-\gamma\nabla L_{gen,k}.
\end{equation}
After several local gradient updates (i.e. $P$ steps), institution $k$ can obtain its local model. However, as discussed earlier, local data not only tend to have a limited capacity, but also may display unique characteristics due to discrepancies in the acquisition protocols. Therefore, the resulting model inevitably suffers from insufficient representation ability and low generalizability to other datasets. Ideally, one can mitigate such issue by training on a diversity-rich multi-source dataset or simply constructing a global dataset $\mathcal{D}$ from all available sources. Nevertheless, due to emerging privacy concerns, it is usually not the case for HFR.
\begin{algorithm}[t!]
	\SetAlgoLined
	\scriptsize
	\textbf{Input:} $\mathcal{D}^1_s, \mathcal{D}^2_s, \dots, \mathcal{D}^K_s $, $K$ dispersed datasets; $P$, local update steps; $Q$, communication rounds; $\gamma$, learning rate; $\Theta_{1},..., \Theta_{K}$, local models; $\Theta$, global model. \\
	$\triangleright$ parameters initialization\\
	\For{q = 0 to Q}{
		\For{k = 0 to K \textbf{in parallel}} {
			$\triangleright$ \text{deploy weights $\Theta^q$ to local model}\\
			\For{p = 0 to P}{
			\textbf{VPGAN Face Hallucination:}\\
			$\triangleright$ compute loss $L_{gen,k}$ using Eq.~\ref{gen}\\
			\textbf{Model Proximity Regularization:}\\
			$\triangleright$ compute the proximal term with respect to $\Theta_{k}^{p,q}$ and $\Theta^q$ \\
			$\triangleright$ compute the final local objective using Eq.~\ref{eq:mpr} and update $\Theta_{k}^{p,q}$
			}
			$\triangleright$ \text{upload weights to the central server}
		}
		$\triangleright$ \text{update the global model using Eq.~\ref{eq:agg}}\\
	}
	\Return{$\Theta^Q$}
	\caption{VPFL with MPR}\label{al:1}
\end{algorithm}

To maximize data utilization and learn a more generic model, we propose a vanilla FL framework based on FedAvg~\cite{fedavg}. Rather than directly sharing private datasets, we leverage a centralized server to indirectly harvest information from all available institutions. This is achieved by periodically aggregating local models and broadcasting the updated results to all participants. A global update in the central server is calculated as follows:
\begin{equation} \label{eq:agg}
\begin{aligned}
\Theta^q = \frac{1}{K}\sum\limits_{k=1}^{K}\Theta_{k}^q,\\
\end{aligned}
\end{equation}
where $q$ represents the $q$-th communication round. The final trained global model $\Theta^Q$ is obtained after $Q$ rounds of client-server communications.
\\
\noindent\textbf{Model Proximity Regularization.}
While our vanilla FL algorithm manages to scale-up training, in real-world applications, the non-i.i.d. data distribution among institutions will still inevitably hurt the performance~\cite{fedprox}. Due to the dissimilar local objectives $L_{gen,k}$ resulting from heterogeneous data distribution, local updates without proper constraints will cause the resulting model to skew toward the optima of its local objective, leading to inconsistency with the global one. Previous works~\cite{flmr,fedada} circumvent this issue via FL adversarial training between source and target domains. However, such methods require directly sharing the latent features between participants, which compromises the privacy-preserving principle.

Inspired by ~\cite{fedprox,afl,feddyn}, here we introduce a new Model Proximity Regularization (MPR) to correct local updates which can be easily combined with VPGAN. As shown in Figure~\ref{fig2}, rather than solely minimizing the local objective $L_{gen,k}$, MPR introduces an extra proximal term for each local solver to force the proximity of latent codes from the current local model and the initial global model. The final local objective $\mathcal{L}_{gen,k}$  is adjusted to
\begin{equation} \label{eq:mpr}
\begin{aligned}
\mathcal{L}_k = L_{gen,k} + \lambda_{d} \bigg(\|w-w^q\|^2+\sum\limits_{i=1}^{n}\|F_{i}-F^{q}_{i}\|^2\bigg) ,\\
\end{aligned}
\end{equation}
where $\lambda_{d}$ is a balancing parameter. In our unified framework VPFL,  we integrate VPGAN as the base model  and use it in the FL framework. VPFL thus can jointly enjoy benefits of strong visual priors and large-scale training for more realistic and faithful hallucination. The detailed training process can be found in Algorithm~\ref{al:1}. 
\section{Experiments}
In our experiments, we focus on synthesizing $512\times512$ VIS faces from $128\times128$ TH images, with an emphasis on both recognition accuracy and image quality. More analysis, discussion and additional results for resolutions less than $128\times128$ can be found in the supplementary material.\\
\noindent {\bf{Dataset and Evaluation Metrics.}}
So far, there are no standardized protocols in this field. Existing methods report results trained and tested on custom datasets/splits. This paper selects two common datasets (VIS-TH~\cite{mallat2018benchmark} and ARL-VTF~\cite{poster2021large}) where high-resolution VIS images are available. To study the effect of data restriction, we intentionally choose one dataset to be more challenging than the other. We create VIS-TH image pairs by cropping $512\times512$ and $128\times128$ face regions respectively. 
\begin{figure}[htp!]
	\centering
	\includegraphics[width=\columnwidth]{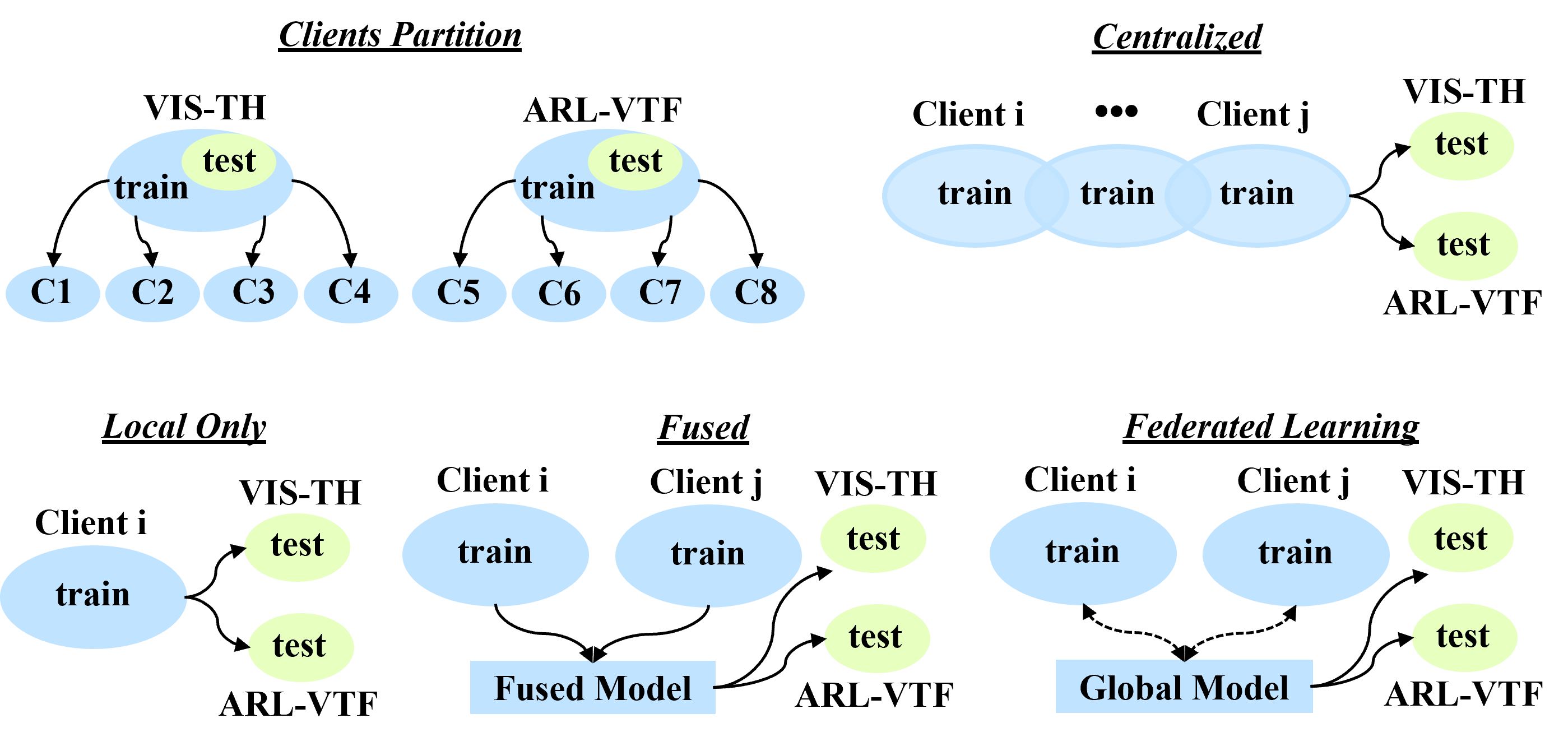}
\vskip-10pt	\caption{The schematics of the clients partition and different training strategies.}\label{fig3}
\vspace{-5mm}
\end{figure}
\begin{figure*}[t!]
	\centering
	\includegraphics[clip, trim=1.8cm 0.60cm 1cm 0cm,width=0.95\textwidth]{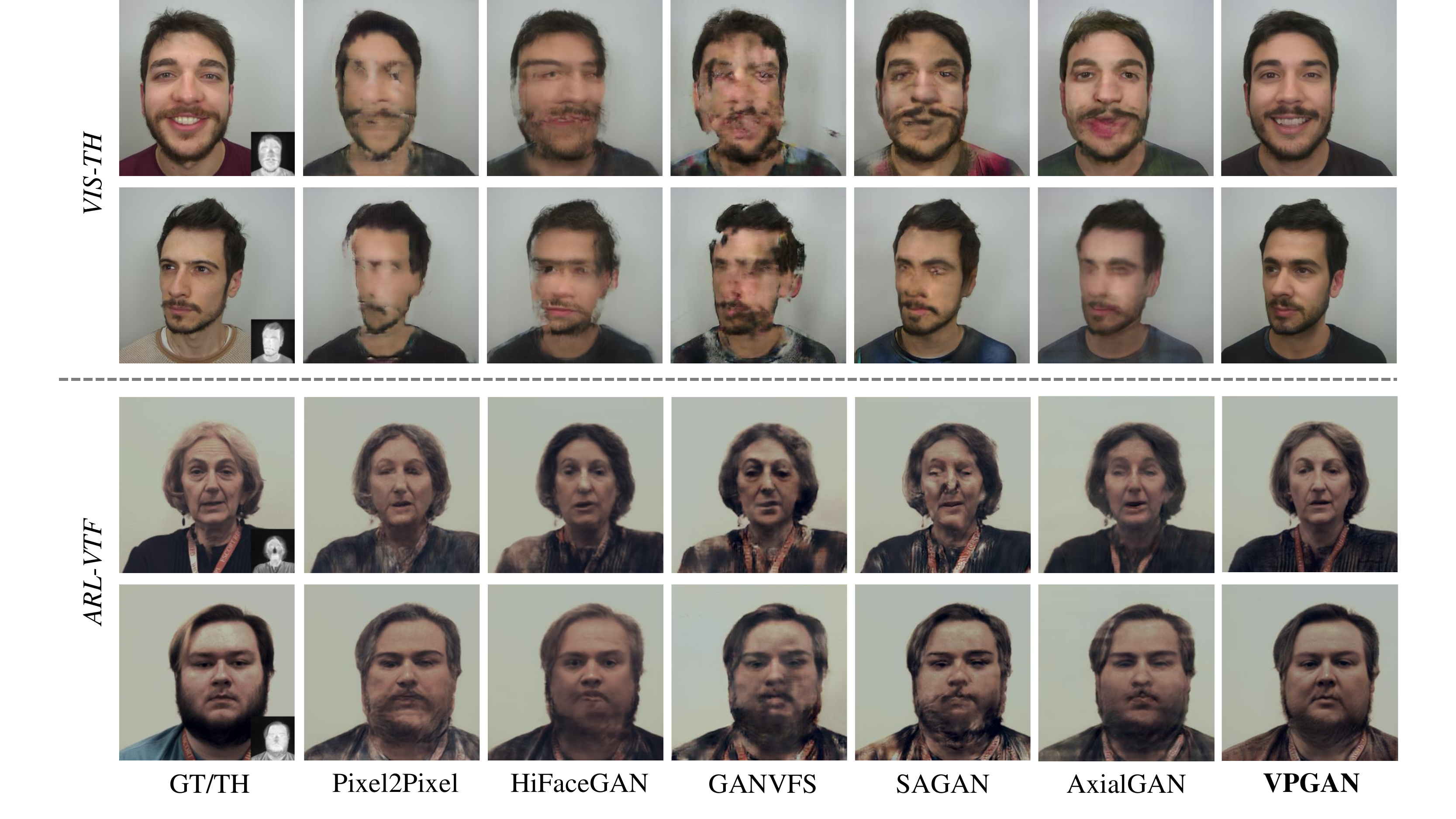}
\vskip-8pt	\caption{Visual comparison on the TH-VIS and ARL-VTF datasets. Low-resolution TH inputs are attached at the  bottom right corner of the GT images with the real scale ratio (128:512) preserved. Our VPGAN can synthesize high-quality faces even with challenging expressions and large poses. Best viewed by zooming
to \textbf{400\%} in the screen.}\label{vpgan_fig}
\vspace{-6mm}
\end{figure*}

\noindent\textbf{VIS-TH} is a challenging dataset containing data from 50 subjects.  Images from each subject contain 21 faces varying significantly in pose, expression and light conditions. VIS-TH images are captured via a dual-sensor camera in LWIR modality and thus naively aligned. We construct the training set by randomly selecting data from 40 subjects. The remaining data from 10 subjects are used as the testing set. 

\noindent\textbf{ARL-VTF} provides subjects' data in LWIR modality with annotations for alignment. We create a  dataset by randomly selecting a subset of 160 subjects with variations only in expressions as the training set, 20 subjects' data as the validation set, and 40 subjects' data as the testing set. The resulting data split contains 3,200 training pairs, 400 validation pairs, and 985 testing pairs. The color adjustment is applied to mitigate overexposure of the VIS images.  

\noindent\textbf{Evaluation Metrics.} This paper extends the existing verification protocol with image quality measurements. For verification, we follow~\cite{duan2020cross} and report Rank-1 accuracy, Verification Rate (VR) @ False Accept Rate (FAR)=$1\%$ and VR@FAR=$0.1\%$. One VIS image of each subject is added to the gallery set and the probe set contains all TH images. 
To measure image quality, we report perceptual metrics LPIPS~\cite{zhang2018unreasonable}, NIQE~\cite{mittal2012making}, identity metric Deg (cosine distance between LightCNN~\cite{wu2018light} features), and pixel-wise PSNR and SSIM~\cite{wang2004image}.

\subsection{Implementation and Training Details}
\noindent \textbf{For VPGAN}, we adopt off-the-shelf StyleGAN2~\cite{karras2020analyzing} as our facial decoder. The UNet encoder contains 5 downsample stages and 7 upsample stages for joint face translation and upsampling. Feature at the lowest level has a spatial size of $4\times 4$. The network is trained using the Adam~\cite{kingma2015adam} optimizer with the following hyperparameters: initial learning rate of 2e-3 for the first 140K iterations then reduced to 1e-3; 150K maximum iterations; batch size of 4; $\lambda_{a}=1$; $\lambda_{b}=10;\lambda_{c}=100$; $\lambda_{d}=$10e-4 if applicable. We implement the proposed model using PyTorch on Nvidia RTX8000 GPUs.

\noindent\textbf{For VPFL}, we adopt the same hyperparameters as that in VPGAN except 80K maximum iterations. The periodical communication between clients and the server is set to 200 iterations. Two training sets are further split into 4 subsets by sampling from a Dirichlet distribution ($\alpha$=0.3) to simulate heterogeneous data distribution in the FL scheme, resulting in 8 independent clients as shown in Figure~\ref{fig3}. Detailed dataset statistics in each client is provided in the supplementary material. For experiments in the FL setting, we not only compare different FL algorithms but also privacy-preserving alternative strategies. Models of \textbf{Local Only} are trained only with data from a single client and evaluated on two testing datasets. We denote the method that obtains independently trained
models from all local clients and fuses their outputs as \textbf{Fused}, which does not violate privacy regulations. In addition, we can obtain a model trained by all available data, which is denoted by \textbf{Centralized}. Since it is prohibited in FL, we treat it as an upper bound. Figure~\ref{fig3} provides the schematics of different training and evaluation strategies in the FL setting.

\begin{table}[h]
       
        \small
        \centering
        \caption{Image quality results on the  \textbf{VIS-TH} dataset.  \redbf{Red} and \blueud{blue} indicates the best and the second best performance.}
         \vspace{-3mm}
        \label{tab:q1}
        \tabcolsep=0.1cm
        \resizebox{\columnwidth}{!}{
                \begin{tabular}{l|cc|c|cc}
                        \hline
                        Methods       & LPIPS$\downarrow$   &NIQE$\downarrow$    & Deg.$\uparrow$   & PSNR$\uparrow$  & SSIM$\uparrow$  \\ \hline \hline
                        TH         &0.7147   &10.666 &36.13 &6.41 &0.3619  \\ \hline
                        Pixel2Pixel~\cite{isola2017image}          &0.3837   &6.642 &43.97 &16.64  &0.6818  \\
                        HiFaceGAN~\cite{yang2020hifacegan}       & 0.3769 &5.973 & 51.03&15.75  & 0.6794 \\
                        GANVFS~\cite{zhang2019synthesis}          & 0.4012 &6.314 &43.95 &16.69  & 0.6569 \\ 
                        SAGAN~\cite{di2019polarimetric}            &  0.2786 & 5.899& 62.35&18.15  &0.7179  \\ 
                        AxialGAN~\cite{immidisetti2021simultaneous}         & \blueud{0.2688}  & \blueud{5.761} &\blueud{62.66}& \redbf{19.02}&\blueud{0.7190}  \\ \hline
                        \textbf{VPGAN (ours)} & \redbf{0.2253}  & \redbf{5.508}& \redbf{68.36}&\blueud{18.96}  &\redbf{0.7456} \\ \hline
                        
        \end{tabular}}
        \vspace{-4mm}
\end{table}

\begin{table}[t] 
    \begin{center}
    \centering
    \caption{Verification results on the \textbf{VIS-TH} dataset.}
   \vspace{-3mm}
   \tabcolsep=0.1cm
    \resizebox{\columnwidth}{!}{
   \label{tab:r1}
    \begin{tabular}{lccc}
        \hline
         Method & Rank-1 & VR@FAR=1\% & VR@FAR=0.1\% \\
        \hline
        LightCNN~\cite{wu2018light}& 30.48 & 8.57& 2.86\\ \hline
        Pixel2Pixel~\cite{isola2017image} &15.24 &2.21 &0.07 \\ 
        HiFaceGAN~\cite{yang2020hifacegan} &44.76  & 10.95& 2.86\\
        GANVFS~\cite{zhang2019synthesis} &18.11 &7.29 &1.90 \\
        SAGAN~\cite{di2019polarimetric}  & 63.33&23.81 &\blueud{17.62} \\
        AxialGAN~\cite{immidisetti2021simultaneous} &\blueud{66.67} &\blueud{24.76} &13.81 \\ \hline
        \textbf{VPGAN (ours)}& \redbf{76.67} & \redbf{45.71}&\redbf{20.00} \\
        \hline
    \end{tabular}}
    \end{center}
   
    \vspace{-4mm}
\end{table}


\subsection{Evaluations for VPGAN}
\noindent {\bf{Results on the VIS-TH dataset.}} 
To demonstrate the effectiveness of our VPGAN, we first report results on the challenging VIS-TH dataset and compare it with 5 representative methods: Pixel2Pixel~\cite{isola2017image}, HiFaceGAN~\cite{yang2020hifacegan}, GANVFS~\cite{zhang2019synthesis}, SAGAN~\cite{di2019polarimetric} and AxialGAN~\cite{immidisetti2021simultaneous}. Pixel2Pixel is a well-known image-to-image translation method. HiFaceGAN is the state-of-the-art approach for face restoration. For TH-VIS hallucination, we select three leading methods GANVFS, SAGAN, and AxialGAN. Note: only AxialGAN has made their code publicly available. For GANVFS and SAGAN, implementations are acquired from authors.

\begin{table}[h]
        \small
        \begin{center}
        \centering
        \caption{Image quality results on the \textbf{ARL-VTF} dataset.}
        \label{tab:q2}
        \vspace{-3mm}
        \tabcolsep=0.1cm
        \resizebox{\columnwidth}{!}{
                \begin{tabular}{l|cc|c|cc}
                        \hline
                        Methods       &
                        LPIPS$\downarrow$   &NIQE$\downarrow$    & Deg.$\uparrow$   & PSNR$\uparrow$  & SSIM$\uparrow$  \\ \hline \hline
                        TH  &0.6721  &10.176&42.34 &5.63 &0.2940  \\ \hline
                        Pixel2Pixel~\cite{isola2017image}          &0.2038 & 6.298&70.67 &19.46 &0.7759  \\
                        HiFaceGAN~\cite{yang2020hifacegan}       &0.2166 &7.274 &70.11 &19.67  &\redbf{0.7954} \\
                        GANVFS~\cite{zhang2019synthesis}       &0.2433 &6.679&67.26 &19.76 & 0.7511 \\ 
                        SAGAN~\cite{di2019polarimetric}            &\blueud{0.1925}  &\blueud{6.155} &\blueud{71.12} &20.11  &0.7772  \\ 
                        AxialGAN~\cite{immidisetti2021simultaneous}          & {0.1998}  & {6.223} &{69.75}& \blueud{20.17}&{0.7770}  \\ \hline
                        \textbf{VPGAN (ours)} & \redbf{0.1713}  & \redbf{6.059}& \redbf{72.00}&\redbf{20.29}  &\blueud{0.7883} \\ \hline
                        
        \end{tabular}}
        \end{center}
        \vspace{-8mm}
\end{table}

\begin{table}[t]
    \begin{center}
    \caption{Verification results on the  \textbf{ARL-VTF} dataset.}
    \vspace{-3mm}
    \tabcolsep=0.1cm
    \resizebox{\columnwidth}{!}{
   \label{tab:r2}
    \begin{tabular}{lccc}
        \hline
         Method & Rank-1 & VR@FAR=1\% & VR@FAR=0.1\% \\
        \hline
        LightCNN~\cite{wu2018light}& 11.07 & 9.24 & 4.57\\ \hline
        Pixel2Pixel~\cite{isola2017image} &70.96 &56.35 &33.60 \\ 
        HiFaceGAN~\cite{yang2020hifacegan} &70.15 &56.65 &32.18 \\
        GANVFS~\cite{zhang2019synthesis} &70.76  &45.99& 22.03\\
        SAGAN~\cite{di2019polarimetric}  &71.16 & 54.11 &\blueud{38.07} \\
        AxialGAN~\cite{immidisetti2021simultaneous} &\blueud{71.57} &\blueud{57.16} &37.36\\ \hline
        \textbf{VPGAN (ours)}& \redbf{74.16} & \redbf{59.96}&\redbf{41.27} \\
        \hline
    \end{tabular}}
    \end{center}
    \vspace{-8mm}
\end{table}

Visual results are shown in Figure \ref{vpgan_fig}. As can be seen from this figure, previous approaches fail to generate clear visible faces. Specifically, Pixel2Pixel, HifaceGAN, GANVFS show strong artifacts and distortions in the generated faces. SAGAN and AxialGAN improve hallucination by adapting self-attention mechanism, but the produced images are still very blurry. In contrast, VPGAN outperforms previous methods by a large margin and it synthesizes the most faithful and accurate faces. Results for quantitative quality assessment are reported in Table \ref{tab:q1}. Our approach achieves the best performance in almost all metrics. VPGAN also obtains the highest Deg. value, indicating its superior ability in preserving identity. All of these results demonstrate the huge benefits of using visual priors for HFR.  

We report verification results in Table \ref{tab:r1}. Given the superior generation quality, there is no surprise that VPGAN achieves the best performance on all metrics. Specifically, our method significantly improves the baseline LightCNN~\cite{wu2018light} by 46\%, and previous state-of-the-art AxialGAN by 10\% in Rank-1 accuracy. In contrast, low-quality hallucinations, e.g., from Pixel2Pixel and GANVFS, can also impair the performance.\\
\noindent {\bf{Results on the ARL-VTF dataset.}} To study the effect of data restriction, we further report results on the ARL-VTF dataset. This dataset contains $\times4$ more subjects (160) than VIS-TH with only slight variations in expressions.  Note that it is non-trivial for a single institution to achieve diversity at this scale. As shown in Figure \ref{vpgan_fig}, data simplicity allows previous methods to yield better visual results as expected. While most of them are able to produce a face outline, they struggle to create detailed facial components. In contrast, our approach can produce realistic and faithful facial details. Its superior hallucination ability is further verified by the quantitative results in Table \ref{tab:q2}. Face verification results are shown in Table \ref{tab:r2}. While previous methods can achieve reasonable performance, VPGAN still reaches the best performance given the more clear and accurate face details. \begin{table}[t]
        
        \small
        \centering
        \caption{Ablation study on the VIS-TH dataset.}
        \vspace{-2mm}
        \label{tab:ab2}
        \tabcolsep=0.1cm
        \resizebox{\columnwidth}{!}{
                \begin{tabular}{c|cc|c|cc}
                        \hline
                               &
                        Rank-1   &VR@FAR=1\%    & Deg.  & LPIPS$\downarrow$  & PSNR$\uparrow$  \\ \hline \hline
                        w/o VP   &52.85 &19.04 &57.69 & 0.2948 &18.23 \\
                        
                        w/o MSCA   &71.90  &31.43 &63.15 &0.2460 &18.74
                        \\ \hline
                        \textbf{VPGAN}          &\redbf{76.19} &\redbf{38.10} &\redbf{65.87} &\redbf{0.2381} &\redbf{18.85}  \\
                        \hline

        \end{tabular}}
        
\end{table}

\subsubsection{Ablation Study}\label{sec:ab}
\noindent\textbf{Effects of Visual Priors.} The key design of VPGAN is to leverage rich and diverse visual priors for better hallucination. Here, we investigate its effectiveness. We construct a baseline model by removing the pre-trained decoder, resulting in a U-Shaped generator. As shown in Table \ref{tab:ab2}, the performance is substantially improved by incorporating visual priors, demonstrating our design is indeed beneficial and helpful. 

\noindent\textbf{Multi-Scale Contexts Aggregation.}
VPGAN improves the standard UNet encoder with Multi-Scale Contexts Aggregation module. Table \ref{tab:ab2} shows its effect. Adding the MSCA can obviously improve results in all metrics, especially in terms of recognition accuracy and Deg. These results indicate that MSCA can provide more accurate generation control and preserve the identity information better. 
\begin{figure}[t!]
	\centering
	\includegraphics[clip, trim=0cm 6.5cm 0.5cm 0cm,width=\columnwidth]{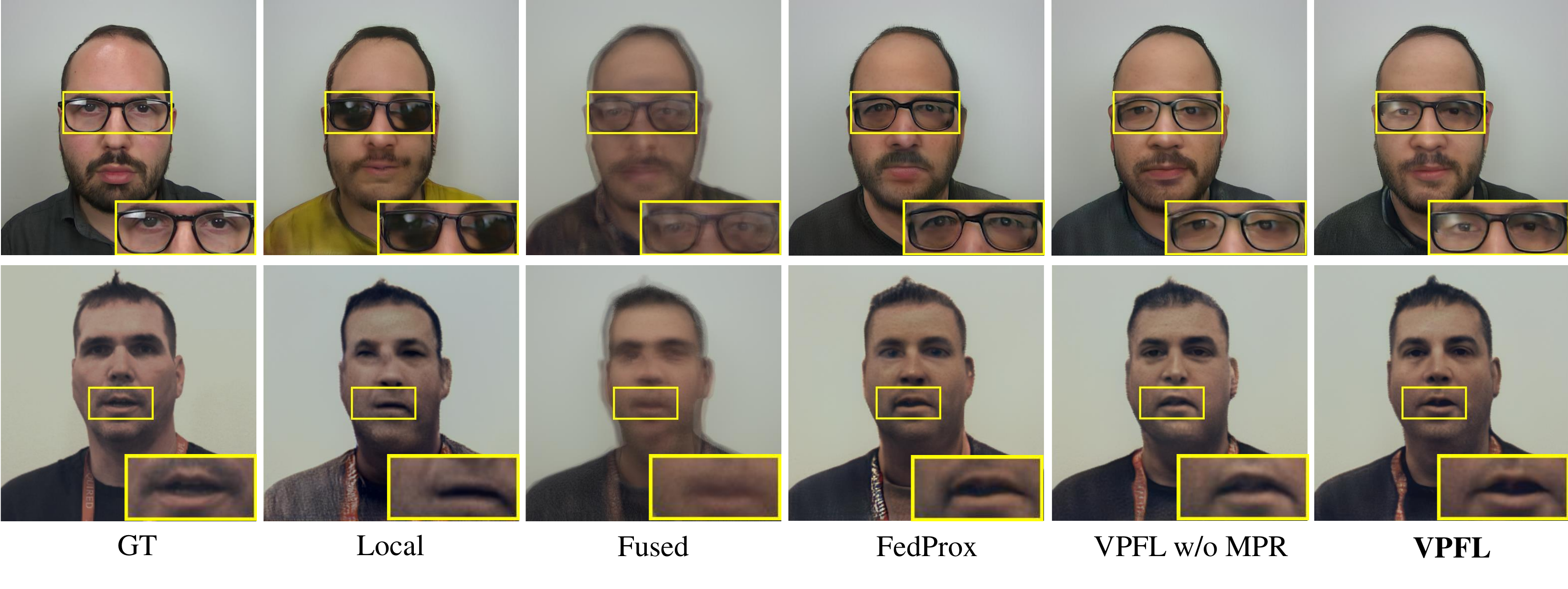}
	\vspace{-5mm}
	\vskip-7pt\caption{Visual comparison under the FL setting. VPFL is able to recover the most accurate face components.}\label{fl_fig}
\vspace{-8mm}
\end{figure}
\begin{table}[]
\setlength{\tabcolsep}{3.5pt}
\centering
\scriptsize
\caption{Verification and image quality comparisons with different methods under the FL setting. VR1\% denotes VR@FAR=1\%.}\label{fl_table}
\vspace{-2mm}
\resizebox{\columnwidth}{!}{
\begin{tabular}{llcccccccc}
\hline
\multicolumn{10}{c}{VIS-TH}                                                                                                                     \\ \hline
\multicolumn{2}{l|}{Methods}       & Rank-1$\uparrow$ & VR1\%$\uparrow$ & \multicolumn{1}{c|}{VR0.1\%$\uparrow$} & \multicolumn{1}{c|}{Deg.$\uparrow$}  & PSNR$\uparrow$  & SSIM$\uparrow$  & LPIPS$\downarrow$  & NIQE$\downarrow$  \\ \hline\hline
\multicolumn{2}{l|}{Local Only C1} & 59.05  & 30.95 & \multicolumn{1}{c|}{17.62}   & \multicolumn{1}{c|}{60.69} & 18.05 & 0.723 & 0.262 & 5.903 \\
\multicolumn{2}{l|}{Local Only C2} & 32.86  & 1.90  & \multicolumn{1}{c|}{0.48}    & \multicolumn{1}{c|}{51.52} & 16.56 & 0.699 & 0.315 & 5.809 \\
\multicolumn{2}{l|}{Local Only C3} & 50.95  & 20.00 & \multicolumn{1}{c|}{7.14}    & \multicolumn{1}{c|}{55.49} & 17.65 & 0.718 & 0.283 & 6.403 \\
\multicolumn{2}{l|}{Local Only C4} & 46.19  & 17.62 & \multicolumn{1}{c|}{10.00}   & \multicolumn{1}{c|}{56.82} & 17.01 & 0.700 & 0.294 & 5.748 \\
\multicolumn{2}{l|}{Local Only C5} & 47.62  & 17.14 & \multicolumn{1}{c|}{8.10}    & \multicolumn{1}{c|}{51.64} & 14.94 & 0.664 & 0.383 & 5.911 \\
\multicolumn{2}{l|}{Local Only C6} & 42.86  & 11.43 & \multicolumn{1}{c|}{6.19}    & \multicolumn{1}{c|}{53.17} & 14.75 & 0.664 & 0.385 & 6.171 \\
\multicolumn{2}{l|}{Local Only C7} & 40.95  & 15.24 & \multicolumn{1}{c|}{5.24}    & \multicolumn{1}{c|}{51.30} & 15.01 & 0.668 & 0.385 & 6.420 \\
\multicolumn{2}{l|}{Local Only C8} & 40.95  & 15.71 & \multicolumn{1}{c|}{5.71}    & \multicolumn{1}{c|}{52.20} & 14.80 & 0.665 & 0.384 & 6.426 \\ \hline\hline
\multicolumn{2}{l|}{Fused}         & 37.62  & 16.19 & \multicolumn{1}{c|}{7.14}    & \multicolumn{1}{c|}{55.83} & 17.36 & \redbf{0.732} & 0.328 & 6.934 \\
\multicolumn{2}{l|}{FedProx~\cite{fedprox}}       & 66.19  & 30.95 & \multicolumn{1}{c|}{20.00}   & \multicolumn{1}{c|}{\blueud{61.99}} & 17.86 & 0.718 & 0.262 & \redbf{5.565} \\
\multicolumn{2}{l|}{VPFL w/o MPR}  & \blueud{70.95}  & \blueud{30.95} & \multicolumn{1}{c|}{\blueud{22.38}}   & \multicolumn{1}{c|}{61.16} & \blueud{18.19} & 0.719 & \blueud{0.254} & \blueud{5.579} \\
\multicolumn{2}{l|}{VPFL}          & \redbf{73.81}  & \redbf{35.71} & \multicolumn{1}{c|}{\redbf{25.71}}   & \multicolumn{1}{c|}{\redbf{65.81}} & \redbf{18.81} & \blueud{0.728} & \redbf{0.245} & 5.651 \\ \hline\hline
\multicolumn{2}{l|}{Centralized}   & 76.67  & 39.05 & \multicolumn{1}{c|}{24.76}   & \multicolumn{1}{c|}{66.63} & 18.71 & 0.743 & 0.232 & 5.729 \\ \hline
\multicolumn{10}{c}{ARL-VTF}                                                                                                                    \\ \hline
\multicolumn{2}{l|}{Methods}      & Rank-1$\uparrow$ & VR1\%$\uparrow$ & \multicolumn{1}{c|}{VR0.1\%$\uparrow$} & \multicolumn{1}{c|}{Deg.$\uparrow$}  & PSNR$\uparrow$  & SSIM$\uparrow$  & LPIPS$\downarrow$  & NIQE$\downarrow$   \\ \hline\hline
\multicolumn{2}{l|}{Local Only C1} & 17.77  & 10.66 & \multicolumn{1}{c|}{2.44}    & \multicolumn{1}{c|}{52.44} & 16.78 & 0.699 & 0.325 & 6.708 \\
\multicolumn{2}{l|}{Local Only C2} & 20.51  & 11.78 & \multicolumn{1}{c|}{2.74}    & \multicolumn{1}{c|}{47.65} & 16.67 & 0.704 & 0.330 & 6.821 \\
\multicolumn{2}{l|}{Local Only C3} & 16.85  & 14.31 & \multicolumn{1}{c|}{2.84}    & \multicolumn{1}{c|}{49.41} & 16.67 & 0.713 & 0.332 & 6.777 \\
\multicolumn{2}{l|}{Local Only C4} & 23.55  & 10.66 & \multicolumn{1}{c|}{3.96}    & \multicolumn{1}{c|}{53.30} & 16.51 & 0.690 & 0.335 & 6.771 \\
\multicolumn{2}{l|}{Local Only C5} & 54.11  & \blueud{39.80} & \multicolumn{1}{c|}{20.10}   & \multicolumn{1}{c|}{64.96} & 19.36 & 0.764 & 0.211 & 6.770 \\
\multicolumn{2}{l|}{Local Only C6} & 54.31  & 36.65 & \multicolumn{1}{c|}{\redbf{22.34}}   & \multicolumn{1}{c|}{65.66} & 19.28 & 0.768 & 0.213 & 6.294 \\
\multicolumn{2}{l|}{Local Only C7} & 40.91  & 33.91 & \multicolumn{1}{c|}{18.07}   & \multicolumn{1}{c|}{63.45} & 19.51 & 0.776 & 0.211 & 6.335 \\
\multicolumn{2}{l|}{Local Only C8} & 54.82  & 37.77 & \multicolumn{1}{c|}{20.91}   & \multicolumn{1}{c|}{64.05} & 19.01 & 0.762 & 0.222 & 6.205 \\ \hline\hline
\multicolumn{2}{l|}{Fused}         & 37.16  & 26.50 & \multicolumn{1}{c|}{10.56}   & \multicolumn{1}{c|}{62.45} & \redbf{19.96} & \redbf{0.789} & 0.260 & 6.840 \\
\multicolumn{2}{l|}{FedProx~\cite{fedprox}}       & 57.77  & 37.36 & \multicolumn{1}{c|}{15.94}   & \multicolumn{1}{c|}{67.21} & 19.60 & 0.770 & 0.212 & 6.022 \\
\multicolumn{2}{l|}{VPFL w/o MPR}  & \blueud{62.03}  & 36.14 & \multicolumn{1}{c|}{16.45}   & \multicolumn{1}{c|}{\blueud{67.46}} & 19.51 & 0.770 & \blueud{0.209} & \blueud{6.019} \\
\multicolumn{2}{l|}{VPFL}          & \redbf{65.79}  & \redbf{40.71} & \multicolumn{1}{c|}{\blueud{22.23}}   & \multicolumn{1}{c|}{\redbf{67.68}} & \blueud{19.69} & \blueud{0.773} & \redbf{0.203} & \redbf{6.013} \\ \hline\hline
\multicolumn{2}{l|}{Centralized}   & 69.34  & 57.77 & \multicolumn{1}{c|}{28.63}   & \multicolumn{1}{c|}{71.28} & 20.08 & 0.785 & 0.186 & 6.106 \\ \hline
\multicolumn{10}{c}{\textbf{Global Test Avg.}}                                                                                                                        \\ \hline
\multicolumn{2}{l|}{Methods}       & Rank-1$\uparrow$ & VR1\%$\uparrow$ & \multicolumn{1}{c|}{VR0.1\%$\uparrow$} & \multicolumn{1}{c|}{Deg.$\uparrow$}  & PSNR$\uparrow$  & SSIM$\uparrow$  & LPIPS$\downarrow$  & NIQE$\downarrow$   \\ \hline\hline
\multicolumn{2}{l|}{Local Only C1} & 38.41  & 20.81 & \multicolumn{1}{c|}{10.03}   & \multicolumn{1}{c|}{56.57} & 17.41 & 0.711 & 0.293 & 6.306 \\
\multicolumn{2}{l|}{Local Only C2} & 26.69  & 6.84  & \multicolumn{1}{c|}{1.61}    & \multicolumn{1}{c|}{49.59} & 16.61 & 0.701 & 0.322 & 6.315 \\
\multicolumn{2}{l|}{Local Only C3} & 33.90  & 17.16 & \multicolumn{1}{c|}{4.99}    & \multicolumn{1}{c|}{52.45} & 17.16 & 0.716 & 0.308 & 6.590 \\
\multicolumn{2}{l|}{Local Only C4} & 34.87  & 14.14 & \multicolumn{1}{c|}{6.98}    & \multicolumn{1}{c|}{55.06} & 16.76 & 0.695 & 0.314 & 6.259 \\
\multicolumn{2}{l|}{Local Only C5} & 50.87  & 28.47 & \multicolumn{1}{c|}{14.10}   & \multicolumn{1}{c|}{58.30} & 17.15 & 0.714 & 0.297 & 6.341 \\
\multicolumn{2}{l|}{Local Only C6} & 48.59  & 24.04 & \multicolumn{1}{c|}{14.27}   & \multicolumn{1}{c|}{59.42} & 17.01 & 0.716 & 0.299 & 6.233 \\
\multicolumn{2}{l|}{Local Only C7} & 40.93  & 24.58 & \multicolumn{1}{c|}{11.66}   & \multicolumn{1}{c|}{57.38} & 17.26 & 0.722 & 0.298 & 6.378 \\
\multicolumn{2}{l|}{Local Only C8} & 47.89  & 26.74 & \multicolumn{1}{c|}{13.31}   & \multicolumn{1}{c|}{58.13} & 16.90 & 0.713 & 0.303 & 6.316 \\ \hline\hline
\multicolumn{2}{l|}{Fused}         & 37.39  & 21.35 & \multicolumn{1}{c|}{8.85}    & \multicolumn{1}{c|}{59.14} & 18.66 & \redbf{0.761} & 0.294 & 6.887 \\
\multicolumn{2}{l|}{FedProx~\cite{fedprox}}       & 61.98  & \blueud{34.16} & \multicolumn{1}{c|}{17.97}   & \multicolumn{1}{c|}{\blueud{64.60}} & 18.73 & 0.744 & 0.237 & \redbf{5.794} \\
\multicolumn{2}{l|}{VPFL w/o MPR}  & \blueud{66.49}  & 33.55 & \multicolumn{1}{c|}{\blueud{19.42}}   & \multicolumn{1}{c|}{64.31} & \blueud{18.85} & 0.745 & \blueud{0.231} & \blueud{5.799} \\
\multicolumn{2}{l|}{VPFL}          & \redbf{69.80}  & \redbf{38.21} & \multicolumn{1}{c|}{\redbf{23.97}}   & \multicolumn{1}{c|}{\redbf{66.75}} & \redbf{19.25} & \blueud{0.751} & \redbf{0.224} & 5.832 \\ \hline\hline
\multicolumn{2}{l|}{Centralized}   & 73.01  & 48.41 & \multicolumn{1}{c|}{26.70}   & \multicolumn{1}{c|}{68.96} & 19.40 & 0.764 & 0.209 & 5.918 \\ \hline
\end{tabular}
}
\vspace{-6mm}
\end{table}

\subsection{Evaluations for VPFL}
Here we show the benefits of collaborative training for HFR by revealing performance and generalizability gaps. Table~\ref{fl_table} presents the verification and quality assessment results of different privacy-preserving strategies on three sub-tables VIS-TH, ARL-VTF, and \emph{Global Test Avg}. \emph{Global Test Avg.} refers to the average performance on the two datasets and reflects generalizability. 

The first 8 rows of each sub-table report the results of locally trained models (\textbf{Local Only}). We treat them as baselines. Due to the data heterogeneity, all locally trained models exhibit low generalizability on the data from another distribution. For example, C1-C4 achieve low performance on ARL-VTF and vice versa. This can also be verified by their poor results on the Global Test Avg. In addition, the naive \textbf{Fused} strategy cannot consistently improve the performance. In contrast, apparent improvements can be observed by introducing the FL scheme, which demonstrates the necessity of collaborative training. When compared with the strong FL baseline FedProx~\cite{fedprox} (with the empirically best $\mu$=10e-4), our VPFL achieves the significantly better overall performance and is closest to the \textbf{Centralized} upper bound. These results indicate that VPFL can generalize well and is more robust towards heterogeneous data distribution. As shown in the last three rows of each sub-table, the advantages come from the newly designed MPR.  These quantitative results are also aligned with the visual comparison in Figure~\ref{fl_fig}. One can see that VPFL yields the most accurate and faithful hallucination results.

\section{Conclusion}
In this paper, we proposed a unified framework VPFL for heterogeneous face hallucination. VPFL consists of a novel VPGAN and a new Federated Learning (FL) scheme. VPGAN introduces powerful visual priors to avoid learning hallucination from scratch, resulting in more accurate generation under the current data limitations. With the consideration of practical privacy issues, the proposed FL scheme allows institution-wise collaborations without sharing data, making large-scale training possible. Extensive experiments demonstrate that VPFL can significantly boost HFR by synthesizing accurate and realistic visible faces at a resolution unseen in the literature. Discussion of limitations can be found in the supplementary material.
\paragraph{Acknowledgments}
This work was supported by NSF CARRER award 2045489.

{\small
\bibliographystyle{ieee_fullname}
\bibliography{egbib}

\begin{thebibliography}{10}\itemsep=-1pt

\bibitem{huang2012buaa}
The buaa-visnir face database instructions.
\newblock 2012.

\bibitem{feddyn}
Durmus Alp~Emre Acar, Yue Zhao, Ramon Matas, Matthew Mattina, Paul Whatmough,
  and Venkatesh Saligrama.
\newblock Federated learning based on dynamic regularization.
\newblock In {\em International Conference on Learning Representations}, 2020.

\bibitem{fedface}
Divyansh Aggarwal, Jiayu Zhou, and Anil~K Jain.
\newblock Fedface: Collaborative learning of face recognition model.
\newblock {\em arXiv preprint arXiv:2104.03008}, 2021.

\bibitem{chan2021glean}
Kelvin~CK Chan, Xintao Wang, Xiangyu Xu, Jinwei Gu, and Chen~Change Loy.
\newblock Glean: Generative latent bank for large-factor image
  super-resolution.
\newblock In {\em IEEE Conference on Computer Vision and Pattern Recognition},
  pages 14245--14254, 2021.

\bibitem{deng2019arcface}
Jiankang Deng, Jia Guo, Niannan Xue, and Stefanos Zafeiriou.
\newblock Arcface: Additive angular margin loss for deep face recognition.
\newblock In {\em IEEE Conference on Computer Vision and Pattern Recognition},
  pages 4690--4699, 2019.

\bibitem{di2019polarimetric}
Xing Di, Benjamin~S Riggan, Shuowen Hu, Nathaniel~J Short, and Vishal~M Patel.
\newblock Polarimetric thermal to visible face verification via self-attention
  guided synthesis.
\newblock In {\em IEEE International Conference on Biometrics}, pages 1--8,
  2019.

\bibitem{di2018polarimetric}
Xing Di, He Zhang, and Vishal~M Patel.
\newblock Polarimetric thermal to visible face verification via attribute
  preserved synthesis.
\newblock In {\em IEEE International Conference on Biometrics Theory,
  Applications and Systems}, pages 1--10, 2018.

\bibitem{duan2020cross}
Boyan Duan, Chaoyou Fu, Yi Li, Xingguang Song, and Ran He.
\newblock Cross-spectral face hallucination via disentangling independent
  factors.
\newblock In {\em IEEE Conference on Computer Vision and Pattern Recognition},
  pages 7930--7938, 2020.

\bibitem{espinosa2013new}
Virginia Espinosa-Dur{\'o}, Marcos Faundez-Zanuy, and Ji{\v{r}}{\'\i} Mekyska.
\newblock A new face database simultaneously acquired in visible, near-infrared
  and thermal spectrums.
\newblock {\em Cognitive Computation}, 5(1):119--135, 2013.

\bibitem{fu2021cm}
Chaoyou Fu, Yibo Hu, Xiang Wu, Hailin Shi, Tao Mei, and Ran He.
\newblock Cm-nas: Cross-modality neural architecture search for
  visible-infrared person re-identification.
\newblock In {\em Proceedings of the IEEE/CVF International Conference on
  Computer Vision}, pages 11823--11832, 2021.

\bibitem{fu2021high}
Chaoyou Fu, Yibo Hu, Xiang Wu, Guoli Wang, Qian Zhang, and Ran He.
\newblock High-fidelity face manipulation with extreme poses and expressions.
\newblock {\em IEEE Transactions on Information Forensics and Security},
  16:2218--2231, 2021.

\bibitem{fu2019dual}
Chaoyou Fu, Xiang Wu, Yibo Hu, Huaibo Huang, and Ran He.
\newblock Dual variational generation for low shot heterogeneous face
  recognition.
\newblock {\em Advances in Neural Information Processing Systems},
  32:2674--2683, 2019.

\bibitem{fu2021dvg}
Chaoyou Fu, Xiang Wu, Yibo Hu, Huaibo Huang, and Ran He.
\newblock Dvg-face: Dual variational generation for heterogeneous face
  recognition.
\newblock {\em IEEE transactions on pattern analysis and machine intelligence},
  2021.

\bibitem{galoogahi2012face}
Hamed~Kiani Galoogahi and Terence Sim.
\newblock Face sketch recognition by local radon binary pattern: Lrbp.
\newblock In {\em IEEE International Conference on Image Processing}, pages
  1837--1840, 2012.

\bibitem{gong2017heterogeneous}
Dihong Gong, Zhifeng Li, Weilin Huang, Xuelong Li, and Dacheng Tao.
\newblock Heterogeneous face recognition: A common encoding feature
  discriminant approach.
\newblock {\em IEEE Transactions on Image Processing}, 26(5):2079--2089, 2017.

\bibitem{goodfellow2014generative}
Ian Goodfellow, Jean Pouget-Abadie, Mehdi Mirza, Bing Xu, David Warde-Farley,
  Sherjil Ozair, Aaron Courville, and Yoshua Bengio.
\newblock Generative adversarial nets.
\newblock {\em Advances in neural information processing systems}, 27, 2014.

\bibitem{gu2020image}
Jinjin Gu, Yujun Shen, and Bolei Zhou.
\newblock Image processing using multi-code gan prior.
\newblock In {\em IEEE Conference on Computer Vision and Pattern Recognition},
  pages 3012--3021, 2020.

\bibitem{flmr}
Pengfei Guo, Puyang Wang, Jinyuan Zhou, Shanshan Jiang, and Vishal~M Patel.
\newblock Multi-institutional collaborations for improving deep learning-based
  magnetic resonance image reconstruction using federated learning.
\newblock In {\em IEEE Conference on Computer Vision and Pattern Recognition},
  pages 2423--2432, 2021.

\bibitem{guo2022auto}
Pengfei Guo, Dong Yang, Ali Hatamizadeh, An Xu, Ziyue Xu, Wenqi Li, Can Zhao,
  Daguang Xu, Stephanie Harmon, Evrim Turkbey, et~al.
\newblock Auto-fedrl: Federated hyperparameter optimization for
  multi-institutional medical image segmentation.
\newblock {\em arXiv preprint arXiv:2203.06338}, 2022.

\bibitem{guo2016ms}
Yandong Guo, Lei Zhang, Yuxiao Hu, Xiaodong He, and Jianfeng Gao.
\newblock Ms-celeb-1m: A dataset and benchmark for large-scale face
  recognition.
\newblock In {\em European Conference on Computer Vision}, pages 87--102.
  Springer, 2016.

\bibitem{he2018wasserstein}
Ran He, Xiang Wu, Zhenan Sun, and Tieniu Tan.
\newblock Wasserstein cnn: Learning invariant features for nir-vis face
  recognition.
\newblock {\em IEEE Transactions on Pattern Analysis and Machine Intelligence},
  41(7):1761--1773, 2018.

\bibitem{huang2017arbitrary}
Xun Huang and Serge Belongie.
\newblock Arbitrary style transfer in real-time with adaptive instance
  normalization.
\newblock In {\em Proceedings of the IEEE international conference on computer
  vision}, pages 1501--1510, 2017.

\bibitem{immidisetti2021simultaneous}
Rakhil Immidisetti, Shuowen Hu, and Vishal~M. Patel.
\newblock Simultaneous face hallucination and translation for thermal to
  visible face verification using axial-gan.
\newblock In {\em IEEE International Joint Conference on Biometrics}, pages
  1--8, 2021.

\bibitem{isola2017image}
Phillip Isola, Jun-Yan Zhu, Tinghui Zhou, and Alexei~A Efros.
\newblock Image-to-image translation with conditional adversarial networks.
\newblock In {\em IEEE Conference on Computer Vision and Pattern Recognition},
  pages 1125--1134, 2017.

\bibitem{karras2019style}
Tero Karras, Samuli Laine, and Timo Aila.
\newblock A style-based generator architecture for generative adversarial
  networks.
\newblock In {\em IEEE Conference on Computer Vision and Pattern Recognition},
  pages 4401--4410, 2019.

\bibitem{karras2020analyzing}
Tero Karras, Samuli Laine, Miika Aittala, Janne Hellsten, Jaakko Lehtinen, and
  Timo Aila.
\newblock Analyzing and improving the image quality of stylegan.
\newblock In {\em IEEE Conference on Computer Vision and Pattern Recognition},
  pages 8110--8119, 2020.

\bibitem{kingma2015adam}
Diederik~P Kingma and Jimmy Ba.
\newblock Adam: A method for stochastic optimization.
\newblock In {\em International Conference on Learning Representations}, 2015.

\bibitem{klare2012heterogeneous}
Brendan~F Klare and Anil~K Jain.
\newblock Heterogeneous face recognition using kernel prototype similarities.
\newblock {\em IEEE Transactions on Pattern Analysis and Machine Intelligence},
  35(6):1410--1422, 2012.

\bibitem{konevcny2016federated}
Jakub Kone{\v{c}}n{\`y}, H~Brendan McMahan, Felix~X Yu, Peter Richt{\'a}rik,
  Ananda~Theertha Suresh, and Dave Bacon.
\newblock Federated learning: Strategies for improving communication
  efficiency.
\newblock {\em arXiv preprint arXiv:1610.05492}, 2016.

\bibitem{lezama2017not}
Jos{\'e} Lezama, Qiang Qiu, and Guillermo Sapiro.
\newblock Not afraid of the dark: Nir-vis face recognition via cross-spectral
  hallucination and low-rank embedding.
\newblock In {\em IEEE Conference on Computer Vision and Pattern Recognition},
  pages 6628--6637, 2017.

\bibitem{li2021federated}
Qinbin Li, Yiqun Diao, Quan Chen, and Bingsheng He.
\newblock Federated learning on non-iid data silos: An experimental study.
\newblock {\em arXiv preprint arXiv:2102.02079}, 2021.

\bibitem{li2007illumination}
Stan~Z Li, RuFeng Chu, ShengCai Liao, and Lun Zhang.
\newblock Illumination invariant face recognition using near-infrared images.
\newblock {\em IEEE Transactions on Pattern Analysis and Machine Intelligence},
  29(4):627--639, 2007.

\bibitem{li2009hfb}
Stan~Z Li, Zhen Lei, and Meng Ao.
\newblock The hfb face database for heterogeneous face biometrics research.
\newblock In {\em IEEE Conference on Computer Vision and Pattern Recognition
  Workshops}, pages 1--8, 2009.

\bibitem{fedprox}
Tian Li, Anit~Kumar Sahu, Manzil Zaheer, Maziar Sanjabi, Ameet Talwalkar, and
  Virginia Smith.
\newblock Federated optimization in heterogeneous networks.
\newblock In {\em Proceedings of Machine Learning and Systems}, volume~2, pages
  429--450, 2020.

\bibitem{li2019convergence}
Xiang Li, Kaixuan Huang, Wenhao Yang, Shusen Wang, and Zhihua Zhang.
\newblock On the convergence of fedavg on non-iid data.
\newblock In {\em International Conference on Learning Representations}, 2019.

\bibitem{fedbn}
Xiaoxiao Li, Meirui JIANG, Xiaofei Zhang, Michael Kamp, and Qi Dou.
\newblock Fed{BN}: Federated learning on non-{IID} features via local batch
  normalization.
\newblock In {\em International Conference on Learning Representations}, 2021.

\bibitem{liao2009heterogeneous}
Shengcai Liao, Dong Yi, Zhen Lei, Rui Qin, and Stan~Z Li.
\newblock Heterogeneous face recognition from local structures of normalized
  appearance.
\newblock In {\em IEEE International Conference on Biometrics}, pages 209--218.
  Springer, 2009.

\bibitem{lin2019face}
Jinpeng Lin, Hao Yang, Dong Chen, Ming Zeng, Fang Wen, and Lu Yuan.
\newblock Face parsing with roi tanh-warping.
\newblock In {\em IEEE Conference on Computer Vision and Pattern Recognition},
  pages 5654--5663, 2019.

\bibitem{liu2017face}
Sifei Liu, Jianping Shi, Ji Liang, and Ming-Hsuan Yang.
\newblock Face parsing via recurrent propagation.
\newblock {\em arXiv preprint arXiv:1708.01936}, 2017.

\bibitem{liu2012heterogeneous}
Sifei Liu, Dong Yi, Zhen Lei, and Stan~Z Li.
\newblock Heterogeneous face image matching using multi-scale features.
\newblock In {\em IEEE International Conference on Biometrics}, pages 79--84,
  2012.

\bibitem{mallat2019cross}
Khawla Mallat, Naser Damer, Fadi Boutros, Arjan Kuijper, and Jean-Luc Dugelay.
\newblock Cross-spectrum thermal to visible face recognition based on cascaded
  image synthesis.
\newblock In {\em IEEE International Conference on Biometrics}, pages 1--8,
  2019.

\bibitem{mallat2018benchmark}
Khawla Mallat and Jean-Luc Dugelay.
\newblock A benchmark database of visible and thermal paired face images across
  multiple variations.
\newblock In {\em International Conference of the Biometrics Special Interest
  Group (BIOSIG)}, pages 1--5, 2018.

\bibitem{fedavg}
Brendan McMahan, Eider Moore, Daniel Ramage, Seth Hampson, and Blaise~Aguera y
  Arcas.
\newblock Communication-efficient learning of deep networks from decentralized
  data.
\newblock In {\em Artificial Intelligence and Statistics}, pages 1273--1282.
  PMLR, 2017.

\bibitem{menon2020pulse}
Sachit Menon, Alexandru Damian, Shijia Hu, Nikhil Ravi, and Cynthia Rudin.
\newblock Pulse: Self-supervised photo upsampling via latent space exploration
  of generative models.
\newblock In {\em IEEE Conference on Computer Vision and Pattern Recognition},
  pages 2437--2445, 2020.

\bibitem{mittal2012making}
Anish Mittal, Rajiv Soundararajan, and Alan~C Bovik.
\newblock Making a “completely blind” image quality analyzer.
\newblock {\em IEEE Signal processing letters}, 20(3):209--212, 2012.

\bibitem{afl}
Mehryar Mohri, Gary Sivek, and Ananda~Theertha Suresh.
\newblock Agnostic federated learning.
\newblock In {\em International Conference on Machine Learning}, pages
  4615--4625. PMLR, 2019.

\bibitem{panetta2018comprehensive}
Karen Panetta, Qianwen Wan, Sos Agaian, Srijith Rajeev, Shreyas Kamath, Rahul
  Rajendran, Shishir~Paramathma Rao, Aleksandra Kaszowska, Holly~A Taylor,
  Arash Samani, et~al.
\newblock A comprehensive database for benchmarking imaging systems.
\newblock {\em IEEE Transactions on Pattern Analysis and Machine Intelligence},
  42(3):509--520, 2018.

\bibitem{park2019semantic}
Taesung Park, Ming-Yu Liu, Ting-Chun Wang, and Jun-Yan Zhu.
\newblock Semantic image synthesis with spatially-adaptive normalization.
\newblock In {\em Proceedings of the IEEE/CVF conference on computer vision and
  pattern recognition}, pages 2337--2346, 2019.

\bibitem{peng2020soft}
Chunlei Peng, Nannan Wang, Jie Li, and Xinbo Gao.
\newblock Soft semantic representation for cross-domain face recognition.
\newblock {\em IEEE Transactions on Information Forensics and Security},
  16:346--360, 2020.

\bibitem{fedada}
Xingchao Peng, Zijun Huang, Yizhe Zhu, and Kate Saenko.
\newblock Federated adversarial domain adaptation.
\newblock {\em arXiv preprint arXiv:1911.02054}, 2019.

\bibitem{HFR_FG21}
N. Peri, J. Gleason, C.~D. Castillo, T. Bourlai, V.~M. Patel, and R. Chellappa.
\newblock A synthesis-based approach for thermal-to-visible face verification.
\newblock In {\em IEEE International Conference on Automatic Face and Gesture
  Recognition}, 2021.

\bibitem{poster2021large}
Domenick Poster, Matthew Thielke, Robert Nguyen, Srinivasan Rajaraman, Xing Di,
  Cedric~Nimpa Fondje, Vishal~M Patel, Nathaniel~J Short, Benjamin~S Riggan,
  Nasser~M Nasrabadi, et~al.
\newblock A large-scale, time-synchronized visible and thermal face dataset.
\newblock In {\em IEEE Winter Conference on Applications of Computer Vision},
  pages 1559--1568, 2021.

\bibitem{Pumarola_2018_ECCV}
Albert Pumarola, Antonio Agudo, Aleix~M. Martinez, Alberto Sanfeliu, and
  Francesc Moreno-Noguer.
\newblock Ganimation: Anatomically-aware facial animation from a single image.
\newblock In {\em European Conference on Computer Vision}. Springer, 2018.

\bibitem{fedopt}
Sashank Reddi, Zachary Charles, Manzil Zaheer, Zachary Garrett, Keith Rush,
  Jakub Kone{\v{c}}n{\`y}, Sanjiv Kumar, and H~Brendan McMahan.
\newblock Adaptive federated optimization.
\newblock {\em arXiv preprint arXiv:2003.00295}, 2020.

\bibitem{richardson2021encoding}
Elad Richardson, Yuval Alaluf, Or Patashnik, Yotam Nitzan, Yaniv Azar, Stav
  Shapiro, and Daniel Cohen-Or.
\newblock Encoding in style: a stylegan encoder for image-to-image translation.
\newblock In {\em IEEE Conference on Computer Vision and Pattern Recognition},
  pages 2287--2296, 2021.

\bibitem{riggan2018thermal}
Benjamin~S Riggan, Nathaniel~J Short, and Shuowen Hu.
\newblock Thermal to visible synthesis of face images using multiple regions.
\newblock In {\em IEEE Winter Conference on Applications of Computer Vision},
  pages 30--38, 2018.

\bibitem{UNet}
Olaf Ronneberger, Philipp Fischer, and Thomas Brox.
\newblock U-net: Convolutional networks for biomedical image segmentation.
\newblock In {\em International Conference on Medical Image Computing and
  Computer-Assisted Intervention}, pages 234--241. Springer, 2015.

\bibitem{schroff2015facenet}
Florian Schroff, Dmitry Kalenichenko, and James Philbin.
\newblock Facenet: A unified embedding for face recognition and clustering.
\newblock In {\em IEEE Conference on Computer Vision and Pattern Recognition},
  pages 815--823, 2015.

\bibitem{shao2014generalized}
Ming Shao, Dmitry Kit, and Yun Fu.
\newblock Generalized transfer subspace learning through low-rank constraint.
\newblock {\em International Journal of Computer Vision}, 109(1-2):74--93,
  2014.

\bibitem{song2018adversarial}
Lingxiao Song, Man Zhang, Xiang Wu, and Ran He.
\newblock Adversarial discriminative heterogeneous face recognition.
\newblock In {\em the AAAI Conference on Artificial Intelligence}, volume~32,
  2018.

\bibitem{wang2018cosface}
Hao Wang, Yitong Wang, Zheng Zhou, Xing Ji, Dihong Gong, Jingchao Zhou, Zhifeng
  Li, and Wei Liu.
\newblock Cosface: Large margin cosine loss for deep face recognition.
\newblock In {\em IEEE Conference on Computer Vision and Pattern Recognition},
  pages 5265--5274, 2018.

\bibitem{wang2021towards}
Xintao Wang, Yu Li, Honglun Zhang, and Ying Shan.
\newblock Towards real-world blind face restoration with generative facial
  prior.
\newblock In {\em IEEE Conference on Computer Vision and Pattern Recognition},
  pages 9168--9178, 2021.

\bibitem{wang2008face}
Xiaogang Wang and Xiaoou Tang.
\newblock Face photo-sketch synthesis and recognition.
\newblock {\em IEEE Transactions on Pattern Analysis and Machine Intelligence},
  31(11):1955--1967, 2008.

\bibitem{wang2018recovering}
Xintao Wang, Ke Yu, Chao Dong, and Chen~Change Loy.
\newblock Recovering realistic texture in image super-resolution by deep
  spatial feature transform.
\newblock In {\em Proceedings of the IEEE conference on computer vision and
  pattern recognition}, pages 606--615, 2018.

\bibitem{wang2004image}
Zhou Wang, Alan~C Bovik, Hamid~R Sheikh, and Eero~P Simoncelli.
\newblock Image quality assessment: from error visibility to structural
  similarity.
\newblock {\em IEEE Transactions on Image Processing}, 13(4):600--612, 2004.

\bibitem{wen2016discriminative}
Yandong Wen, Kaipeng Zhang, Zhifeng Li, and Yu Qiao.
\newblock A discriminative feature learning approach for deep face recognition.
\newblock In {\em European Conference on Computer Vision}, pages 499--515.
  Springer, 2016.

\bibitem{wu2018light}
Xiang Wu, Ran He, Zhenan Sun, and Tieniu Tan.
\newblock A light cnn for deep face representation with noisy labels.
\newblock {\em IEEE Transactions on Information Forensics and Security},
  13(11):2884--2896, 2018.

\bibitem{xia2021gan}
Weihao Xia, Yulun Zhang, Yujiu Yang, Jing-Hao Xue, Bolei Zhou, and Ming-Hsuan
  Yang.
\newblock Gan inversion: A survey.
\newblock {\em arXiv preprint arXiv:2101.05278}, 2021.

\bibitem{xu2022closing}
An Xu, Wenqi Li, Pengfei Guo, Dong Yang, Holger Roth, Ali Hatamizadeh, Can
  Zhao, Daguang Xu, Heng Huang, and Ziyue Xu.
\newblock Closing the generalization gap of cross-silo federated medical image
  segmentation.
\newblock {\em arXiv preprint arXiv:2203.10144}, 2022.

\bibitem{yang2020hifacegan}
Lingbo Yang, Shanshe Wang, Siwei Ma, Wen Gao, Chang Liu, Pan Wang, and Peiran
  Ren.
\newblock Hifacegan: Face renovation via collaborative suppression and
  replenishment.
\newblock In {\em ACM International Conference on Multimedia}, pages
  1551--1560, 2020.

\bibitem{yang2021gan}
Tao Yang, Peiran Ren, Xuansong Xie, and Lei Zhang.
\newblock Gan prior embedded network for blind face restoration in the wild.
\newblock In {\em IEEE Conference on Computer Vision and Pattern Recognition},
  pages 672--681, 2021.

\bibitem{yi2007face}
Dong Yi, Rong Liu, RuFeng Chu, Zhen Lei, and Stan~Z Li.
\newblock Face matching between near infrared and visible light images.
\newblock In {\em IEEE International Conference on Biometrics}, pages 523--530.
  Springer, 2007.

\bibitem{ijcai2019-143}
Junchi Yu, Jie Cao, Yi Li, Xiaofei Jia, and Ran He.
\newblock Pose-preserving cross spectral face hallucination.
\newblock In {\em International Joint Conference on Artificial Intelligence},
  pages 1018--1024, 7 2019.

\bibitem{zeng2021sketchedit}
Yu Zeng, Zhe Lin, and Vishal~M Patel.
\newblock Sketchedit: Mask-free local image manipulation with partial sketches.
\newblock {\em arXiv preprint arXiv:2111.15078}, 2021.

\bibitem{zhang2017generative}
He Zhang, Vishal~M Patel, Benjamin~S Riggan, and Shuowen Hu.
\newblock Generative adversarial network-based synthesis of visible faces from
  polarimetrie thermal faces.
\newblock In {\em IEEE International Joint Conference on Biometrics}, pages
  100--107, 2017.

\bibitem{zhang2019synthesis}
He Zhang, Benjamin~S Riggan, Shuowen Hu, Nathaniel~J Short, and Vishal~M Patel.
\newblock Synthesis of high-quality visible faces from polarimetric thermal
  faces using generative adversarial networks.
\newblock {\em International Journal of Computer Vision}, 127(6):845--862,
  2019.

\bibitem{zhang2018unreasonable}
Richard Zhang, Phillip Isola, Alexei~A Efros, Eli Shechtman, and Oliver Wang.
\newblock The unreasonable effectiveness of deep features as a perceptual
  metric.
\newblock In {\em IEEE Conference on Computer Vision and Pattern Recognition},
  pages 586--595, 2018.

\end{thebibliography}
}

\end{document}